# Functional Principal Component Analysis and Randomized Sparse Clustering Algorithm for Medical Image Analysis

Nan Lin[1], Junhai Jiang[1], Shicheng Guo[1,2] and Momiao Xiong[1,*]

[1]Human Genetics Center, Division of Biostatistics, School of Public Health, The University of Texas Health Science Center at Houston, Houston, TX 77030, USA

[2]State Key Laboratory of Genetic Engineering and Ministry of Education Key Laboratory of Contemporary Anthropology, Collaborative Innovation Center for Genetics and Development, School of Life Sciences and Institutes of Biomedical Sciences, Fudan University, Shanghai 200433, China

[*]Address for correspondence and reprints: Dr. Momiao Xiong, Division of Biostatistics, Human Genetics Center, The University of Texas Health Science Center at Houston, 1200 Pressler st, Houston, Texas 77030, (Phone): 713-500-9894, (Fax): 713-500-0900, E-mail: Momiao.Xiong@uth.tmc.edu

**ABSTRACT**

Due to advances in sensors, growing large and complex medical image data have the ability to visualize the pathological change in the cellular or even the molecular level or anatomical changes in tissues and organs. As a consequence, the medical images have the potential to enhance diagnosis of disease, prediction of clinical outcomes, characterization of disease progression, management of health care and development of treatments, but also pose great methodological and computational challenges for representation and selection of features in image cluster analysis. To address these challenges, we first extend one dimensional functional principal component analysis to the two dimensional functional principle component analyses (2DFPCA) to fully capture space variation of image signals. Image signals contain a large number of redundant and irrelevant features which provide no additional or no useful information for cluster analysis. Widely used methods for removing redundant and irrelevant features are sparse clustering algorithms using a lasso-type penalty to select the features. However, the accuracy of clustering using a lasso-type penalty depends on how to select penalty parameters and a threshold for selecting features. In practice, they are difficult to determine. Recently, randomized algorithms have received a great deal of attention in big data analysis. This paper presents a randomized algorithm for accurate feature selection in image cluster analysis. The proposed method is applied to ovarian and kidney cancer histology image data from the TCGA database. The results demonstrate that the randomized feature selection method coupled with functional principal component analysis substantially outperforms the current sparse clustering algorithms in image cluster analysis.

**Highlights:**

We develop a novel two dimensional FPCA method for image data reduction.

Using feature selection we develop sparse clustering algorithms.

We use provable randomized algorithm to select features for cluster analysis.

We develop a novel randomized sparse clustering algorithm for imaging data analysis.

## *1. Introduction*

Image clustering is to cluster the objects into groups such that the objects within the same group are similar, while the objects in different groups are dissimilar (Yang et al. 2010; Bong and Rajeswari, 2012). Image clustering is a powerful tool to better organize and represent images in image annotation, image indexing, and segmentation and subtype disease identification. Data dimension reduction is essential to the success of image clustering analysis.

Feature extraction and feature selection are two popular types of methods for dimensionality reduction. A widely used method for feature extraction is principal component analysis (PCA). However, PCA does not explore spatial information. It takes the set of spectral images as an unordered set of high dimensional pixels (Gupta and Jacobson, 2006). Spatial information is very important for image cluster and classification analysis. To overcome limitations of PCA and to utilize spatial information of the image signal, the functional expansion of images based on Fourier and wavelet transform are proposed as a useful tool for image feature extraction and data denoising (Strela et al. 1999). Recently, wavelet PCA in which we compute principal components for a set of wavelet coefficients is proposed (Gupta and Jacobson, 2006) to explore both spatial and spectral information. The wavelet PCA improves efficiency to extract image features, but not explicitly considers smoothing image signals over space. To overcome this limitation and fully utilize both spatial and spectral information, we extend one dimensional functional principal component analysis (FPCA) to two dimensional FPCA.

Traditional statistical methods for image cluster and classification analysis often fail to obtain accurate results because of the high dimensional nature of image data (Samiappan et al. 2013). Noisy and irrelevant features result in overfitting. The high dimensionality makes the clustering algorithms very slow (Boutsidis and Magdon-Ismail, 2013). The high dimensionality of image provides a considerable challenge for designing efficient clustering algorithms (Boutsidis et al.

2013). Removing noise, redundant and irrelevant features and retaining a minimal feature subset will dramatically improve the accuracy of image cluster analysis (Aroquiaraj and Thangavel, 2013). The sparse method is a widely used method for feature selection in which a lasso-type penalty provides a general framework to simultaneously find the clusters and the important clustering features in image cluster analysis (Witten and Tibshirani, 2010; Kondo et al. 2012). Although the sparse clustering methods can improve accuracy, it may fail to generate reasonable clusters when the data include a few outliers. In practice, the performance of sparse clustering depends on the selection of penalty parameters and threshold for cutting off features. However, the selection of penalty parameters and threshold proves difficult. Despite the success of feature selection in image clustering, very few provable accurate feature selection methods for clustering exist (Boutsidis et al. 2013).

Alternatively, a randomized method is proving useful when the number of features is prohibitively large (Stracuzzi 2008). An efficient randomized feature selection method for $k$ - means clustering randomly selects the features with probabilities that are calculated via singular value decomposition of the data matrix (Boutsidis and Magdon-Ismail, 2013; Boutsidis et al. 2013). This algorithm has a very useful property that can theoretically guarantee the quality of the clusters. To the best of our knowledge, this efficient and provable accurate randomized feature selection algorithm has not been applied to the image cluster analysis.

Although feature selection and feature extraction are widely used to reduce the dimensionality of the image, we have observed very few practices to combine feature selection and feature extraction together for dimension reduction. We can expect that applying feature selection algorithm to select extracted features from a set of artificial features that are computed via feature extraction will improve the accuracy of image clustering.

The purpose of this paper is to develop a comprehensive sparse clustering algorithm with four components for image cluster analysis. The first component is to use high dimensional FPCA as a feature extraction technique. The second component includes a theoretically provable accurate randomized feature selection algorithm. The third component is to combine feature selection and feature extraction together for dimensionality reduction. The fourth component is spectral clustering with low rank matrix decomposition that can effectively remove noises and ensure the robustness of the algorithms. To evaluate its performance for image cluster analysis, the proposed method is applied to 176 ovarian cancer histology images with the drug response status (106 images with positive drug response and 70 images with drug resistance) and 188 kidney histology images (121 images from tumor samples and 67 images from normal samples) from the TCGA database. Our results strongly demonstrate that the proposed method for feature selection substantially outperform other existing feature selection methods. Software for implementing the proposed methods can be downloaded from our website

http://www.sph.uth.tmc.edu/hgc/faculty/xiong/index.htm and http://www.bioconductor.org/.

## 2. Material and methods

### 2.1. Two dimensional functional principal component analysis

One dimensional functional principal component analysis (FPCA) has been well developed (Ramsay and Silverman, 2005). Now we extend one dimensional FPCA to two dimensional FPCA. Consider a two dimensional region. Let $s$ and $t$ denote coordinates in the $s$ axis and $t$ axis, respectively. Let $x(s,t)$ be a centered image signal located at $s$ and $t$ of the region. The signal $x(s,t)$ is a function of locations $s$ and $t$.

Consider a linear combination of functional values:

$$f = \iint_{S\,T} \beta(s,t)x(s,t)dsdt,$$

where $\beta(s,t)$ is a weight function. To capture the variations in the random functions, we chose weight function $\beta(s,t)$ to maximize the variance of $f$. By the formula for the variance of stochastic integral (Henderson and Plaschko, 2006), we have

$$\text{var}(f) = \iiiint_{S\,T\,S\,T} \beta(s_1,t_1)R(s_1,t_1,s_2,t_2)\beta(s_2,t_2)ds_1dt_1ds_2t_2 \,, \qquad (1)$$

where $R(s_1,t_1,s_2,t_2) = \text{cov}(x(s_1,t_1), x(s_2,t_2))$ is the covariance function of the image signal function $x(s,t)$. Since multiplying $\beta(t)$ by a constant will not change the maximizer of the variance $Var(f)$, we impose a constraint to make the solution unique:

$$\iint_{T\,T} \beta^2(s,t)dsdt = 1. \qquad (2)$$

Therefore, to find the weight function, we seek to solve the following optimization problem:

$$\begin{aligned}\max \quad & \iiiint_{S\,T\,S\,T} \beta(s_1,t_1)R(s_1,t_1,s_2,t_2)\beta(s_2,t_2)ds_1dt_1ds_2t_2 \\ \text{s.t.} \quad & \iint_{S\,T} \beta^2(s,t)dsdt = 1.\end{aligned} \qquad (3)$$

Using the Lagrange multiplier, we reformulate the constrained optimization problem (3) into the following non-constrained optimization problem:

$$\max_{\beta} \quad \frac{1}{2}\iiiint_{S\,T\,S\,T} \beta(s_1,t_1)R(s_1,t_1,s_2,t_2)\beta(s_2,t_2)ds_1dt_1ds_2t_2 + \frac{1}{2}\lambda(1 - \iint_{S\,T}\beta^2(s_1,t_1)ds_1dt_1), \quad (4)$$

where $\lambda$ is a penalty parameter.

By variation calculus (Sagan, 1969), we define the functional

$$J[\beta] = \frac{1}{2}\iiiint_{S\,T\,S\,T} \beta(s_1,t_1)R(s_1,t_1,s_2,t_2)\beta(s_2,t_2)ds_1dt_1ds_2t_2 + \frac{1}{2}\lambda(1 - \iint_{S\,T}\beta^2(s_1,t_1)ds_1dt_1).$$ Its first variation is given by

$$\begin{aligned}
\delta J[h] &= \frac{d}{d\varepsilon} J[\beta(s,t) + \varepsilon h(s,t)] \\
&= \int_S \int_T [\int_S \int_T [R(s_1,t_1,s_2,t_2)\beta(s_2,t_2)ds_2 t_2 - \lambda \beta(s_1,t_1)]h(s_1,t_1)ds_1 dt_1 \\
&= \int_S \int_T [\int_S \int_T R(s_1,t_1,s_2,t_2)\beta(s_2,t_2)ds_2 t_2 - \lambda \beta(s_1,t_1)]^2 ds_1 dt_1 = 0,
\end{aligned}$$

which implies the following integral equation

$$\int_S \int_T R(s_1,t_1,s_2,t_2)\beta(s_2,t_2)ds_2 dt_2 = \lambda \beta(s_1,t_1) \tag{5}$$

for an appropriate eigenvalue $\lambda$. The left side of the integral equation (5) defines a two dimensional integral transform $R$ of the weight function $\beta$. Therefore, the integral transform of the covariance function $R(s_1,t_1,s_2,t_2)$ is referred to as the covariance operator $R$. The integral equation (5) can be rewritten as

$$R\beta = \lambda \beta, \tag{6}$$

where $\beta(s,t)$ is an eigenfunction and referred to as a principal component function. Equation (6) is also referred to as a two dimensional eigenequation. Clearly, the eigenequation (6) looks the same as the eigenequation for the multivariate PCA if the covariance operator and eigenfunction are replaced by the covariance matrix and eigenvector.

Since the number of function values is theoretically infinite, we may have an infinite number of eigenvalues. Provided the functions $X_i$ and $Y_i$ are not linearly dependent, there will be only $N-1$ nonzero eigenvalues, where $N$ is the total number of is sampled individuals ($N = n_A + n_G$, $n_A$ is the number of individuals sampled from cases and $n_G$ is the number of individuals sampled from controls). Eigenfunctions satisfying the eigenequation are orthonormal (Ramsay and Silverman, 2005). In other words, equation (6) generates a set of principal component functions:

$$R\beta_k = \lambda_k \beta_k, \qquad \text{with } \lambda_1 \geq \lambda_2 \geq \cdots.$$

These principal component functions satisfy

(1) $\iint_{S\,T} \beta_k^2(s,t)\,ds\,dt = 1$ and

(2) $\iint_{S\,T} \beta_k(s,t)\beta_m(s,t)\,ds\,dt = 0,$ for all $m < k$.

The principal component function $\beta_1$ with the largest eigenvalue is referred to as the first principal component function and the principal component function $\beta_2$ with the second largest eigenvalue is referred to as the second principal component function, and continuing.

## *2.2. Computations for the principal component function and the principal component score*

The eigenfunction is an integral function and difficult to solve in closed form. A general strategy for solving the eigenfunction problem in (5) is to convert the continuous eigen-analysis problem to an appropriate discrete eigen-analysis task (Ramsay and Silverman 2005). In this paper, we use basis function expansion methods to achieve this conversion.

Let $\{\phi_j(t)\}$ be the series of Fourier functions. For each j, define $\omega_{2j-1} = \omega_{2j} = 2\pi j$. We expand each image signal function $x_i(s,t)$ as a linear combination of the basis function $\phi_j$:

$$x_i(s,t) = \sum_{k=1}^{K}\sum_{l=1}^{K} c_{kj}^{(i)} \phi_k(s)\phi_l(t). \qquad (7)$$

Let t $C_i = [c_{11}^{(i)},...,c_{1K}^{(i)},c_{21}^{(i)},...,c_{2K}^{(i)},...,c_{K1}^{(i)},...,c_{KK}^{(i)}]^T$ and $\phi(t) = [\phi_1(t),\cdots,\phi_K(t)]^T$. Then, equation (7) can be rewritten as

$$x_i(s,t) = C_i^T (\phi(s) \otimes \phi(t)),$$

where $\otimes$ denotes the Kronecker product of two matrices.

Define the vector-valued function $X(s,t) = [x_1(s,t), \cdots, x_N(s,t)]^T$. The joint expansion of all N random functions can be expressed as

$$X(s,t) = C(\phi(s) \otimes \phi(t)) \tag{8}$$

where the matrix C is given by

$$C = \begin{bmatrix} C_1^T \\ \vdots \\ C_N^T \end{bmatrix}.$$

In matrix form the variance-covariance function of the image signal function can be expressed as

$$\begin{aligned} R(s_1,t_1,s_2,t_2) &= \frac{1}{N} X^T(s_1,t_1) X(s_2,t_2) \\ &= \frac{1}{N} [\phi^T(s_1) \otimes \phi^T(t_1)] C^T C [\phi(s_2) \otimes \phi(t_2)]. \end{aligned} \tag{9}$$

Similarly, the eigenfunction $\beta(s,t)$ can be expanded as

$$\beta(s,t) = \sum_{j=1}^{K} \sum_{k=1}^{K} b_{jk} \phi_j(s) \phi_k(t) \text{ or}$$

$$\beta(s,t) = [\phi^T(s) \otimes \phi^T(t)] b, \tag{10}$$

where $b = [b_{11}, ..., b_{1K}, ..., b_{K1}, ..., b_{KK}]^T$. Substituting expansions (9) and (10) of the variance-covariance $R(s_1,t_1,s_2,t_2)$ and eigenfunction $\beta(s,t)$ into the functional eigenequation (5), we obtain

$$[\phi^T(s_1) \otimes \phi^T(t_1)] \frac{1}{N} C^T C b = \lambda [\phi^T(s_1) \otimes \phi^T(t_1)] b. \tag{11}$$

Since equation (11) must hold for all $s$ and $t$, we obtain the following eigenequation:

$$\frac{1}{N} C^T C b = \lambda b. \tag{12}$$

Solving eigenequation (12), we obtain a set of orthonormal eigenvectors $b_j$. A set of orthonormal eigenfunctions is given by

$$\beta_j(s,t) = [\phi^T(s) \otimes \phi^T(t)]b_j, j=1,...,J. \tag{13}$$

The random functions $x_i(s,t)$ can be expanded in terms of eigenfunctions as

$$x_i(t,s) = \sum_{j=1}^{J} \xi_{ij}\beta_j(s,t), i=1,...,N, \tag{14}$$

where $\xi_{ij} = \int_S \int_T x_i(t,s)\beta_j(s,t)dsdt.$, $i=1,...,N, j=1,...,J$ are FPC scores.

## *2.3. Randomized feature selection for $k$ - means clustering*

The most widely used clustering method in practice is $k$-means algorithm. However, using $k$ means to cluster millions or billions of features is not simple and straightforward (Boutsidis and Mardon-Ismail, 2013). An attractive strategy is to select a subset of features and optimize the $k$-means of objective on the low dimensional representation of the original high dimensional data. A natural question is whether feature selection will lose valuable information by throwing away potentially useful features which could lead to a significantly higher clustering error. Here, we introduce a randomized feature selection algorithm with provable guarantees (Boutsidis et al. 2013).

For the self-contain, we begin with a linear algebraic formulation of $k$-means algorithm. Many materials are from Boutsidis et al. 2013. Consider a set of $m$ points:

$$A^T = [P_1,...,P_m] \in R^{n \times m}.$$

A $k$ partition of $m$ points is a collection

$$S = \{S_1, S_2,...,S_k\},$$

of $k$ non-empty pairwise disjoint sets covering the entire dataset. K-means clustering minimizes the *within-cluster sum of squares*. Let $s_j = |S_j|$, be the size of $S_j$. For each set $S_j$, define its centroid (the mean of data points within the set $S_j$):

$$\mu_j = \frac{1}{s_j} \sum_{P_i \in S_j} P_i.$$

The $k$-means objective function is defined as

$$F(P,S) = \sum_{i=1}^{m} \| P_i - \mu(P_i) \|_2^2, \quad (15)$$

where $\mu(P_i)$ is the centroid of the cluster to which $P_i$ belongs.

The $k$-means objective function can be transformed to a more convenient linear algebraic formulation. A $k$ clustering $S$ of $A$ can be represented by its clustering indicator matrix $X \in R^{n \times m}$. Specifically, its element $X_{ij}$ is defined as

$$X_{ij} = \begin{cases} \frac{1}{\sqrt{s_j}} & P_i \in S_j \\ 0 & \text{otherwise.} \end{cases}$$

Each row of $X$ has one non-zero element, corresponding to the cluster to which the data point belongs. Each column has $s_j$ non-zero elements, which denotes the data points belonging to cluster $S_j$. The linear algebraic formulation of the $k$-means objective function can be expressed as

$$\begin{aligned} F(A,X) &= \| A - XX^T A \|_F^2 \\ &= \sum_{i=1}^{m} \| P_i^T - X_i X^T A \|_2^2 \quad (16) \\ &= \sum_{i=1}^{m} \| P_i^T - \mu(P_i)^T \|_2^2, \end{aligned}$$

where $\|W\|_F = \sqrt{Tr(W^T W)}$ is the Frobenius norm of a matrix $W$, $X_i$ is the $i$ th row of $X$,

$X^T A = [\mu_1^T, ..., \mu_k^T]^T$ and $X_i X^T X = \mu(P_i)^T$.

Our goal is to find an indicator matrix $X_{opt}$ which minimizes $\|A - XX^T A\|_F^2$:

$$X_{opt} = \underset{X \in R^{m \times k}}{\arg\min} \|A - XX^T A\|_F^2.$$

Define

$$F_{opt} = \|A - X_{opt} X_{opt}^T A\|_F^2.$$

It is noted that $X_{opt} X_{opt}^T A$ has rank at most $k$. Singular value decomposition of the matrix $A$ is given by

$$A = U_k \Sigma_k V_k^T + U_{\rho-k} \Sigma_{\rho-k} V_{\rho-k}^T,$$

where $\rho \leq \min(m,n)$ is rank of the matrix $A$, $U_k \in R^{m \times k}$ and $U_{\rho-k} \in R^{m \times (\rho-k)}$ contain the left singular vectors of $A$, $V_k \in R^{n \times k}$ and $V_{\rho-k} \in R^{n \times (\rho-k)}$ contain the right singular vectors. Singular values $\sigma_1 \geq \sigma_2 \geq ... \geq \sigma_\rho > 0$ are contained in the matrices $\Sigma_k \in R^{k \times k}$ and $\Sigma_{\rho-k} \in R^{(\rho-k)(\rho-k)}$. Let

$A_k = U_k \Sigma_k V_k^T = AV_k V_k^T$ and $A_{\rho-k} = U_{\rho-k} \Sigma_{\rho-k} V_{\rho-k}^T = A - A_k$. Since $A_k$ is the best rank $k$ approximation to $A$ and $X_{opt} X_{opt}^T A$ has rank at most $k$, we obtain

$$\|A - A_k\|_F^2 \leq \|A - X_{opt} X_{opt}^T A\|_F^2 \leq F_{opt}.$$

The feature selection for $k$-means clustering algorithm is to select a subset of $r$ columns $C \in R^{m \times r}$ from $A$, which is a representation of the $m$ data points in the low $r$-dimensional selected feature space. Then, the goal of $k$-means clustering algorithm in the selected feature space is to find partition of $m$ which minimizes $\|C - XX^T C\|_F^2$:

$$\tilde{X}_{opt} = \underset{X \in R^{m \times k}}{\arg\min} \|C - XX^T C\|_F^2.$$

Therefore, feature selection is to seek selection of features such that

$$\|A - \tilde{X}_{opt}\tilde{X}_{opt}^T A\|_F^2 \leq \gamma \|A - X_{opt}X_{opt}^T A\|_F^2. \tag{17}$$

The basic idea of randomized feature selection is that any matrix $C$ which can be used to approximate matrix $A$ can also be used for dimensionality reduction in $k$-means cluster analysis (Boutsidis et al. 2013; Drineas et al. 1999). We seek matrix $C$ that minimizes

$$\begin{aligned}\|A - \tilde{X}_{opt}\tilde{X}_{opt}^T A\|_F^2 &= \|A_k - \tilde{X}_{opt}\tilde{X}_{opt}^T A_k\|_F^2 + \|A_{p-k} - \tilde{X}_{opt}\tilde{X}_{opt}^T A_{p-k}\|_F^2 \\ &= \|AV_k V_k^T - \tilde{X}_{opt}\tilde{X}_{opt}^T AV_k V_k^T\|_F^2 + \|A_{p-k} - \tilde{X}_{opt}\tilde{X}_{opt}^T A_{p-k}\|_F^2 \\ &= \|AV_k - \tilde{X}_{opt}\tilde{X}_{opt}^T AV_k\|_F^2 + \|A_{p-k} - \tilde{X}_{opt}\tilde{X}_{opt}^T A_{p-k}\|_F^2\end{aligned} \tag{18}$$

Let $C = AV_k$. Then, $A_k = CV_k^T$. The minimization problem (18) can be reduced to minimizing $\|C - XX^T C\|_F^2$.

Using formula $C = AV_k$ to calculate the matrix $C$ requires the entire dataset $A$. Next, our goal is to select columns of the matrix $A$ to approximate $C$. We denote the sampling matrix $\Omega = [e_{i_1}, ..., e_{i_r}] \in R^{n \times r}$, where $e_i$ are the standard basis vectors with its $i$th element being one and all other elements being zeroes. Let $S \in R^{r \times r}$ be a diagonal rescaling matrix. Define $C = A\Omega S$. The matrices $\Omega$ and $S$ can be generated by randomized sampling. Since singular value decomposition of a large matrix $A$ may be difficult, we will also use a sampling algorithm to generalize a matrix $Z$ which approximates $V_k$. Thus, the matrix $A$ can be decomposed to $A = AZZ^T + E$, where the matrix $E \in R^{m \times n}$. We still use $\tilde{X}_{opt}$ to denote the output cluster indicator matrix of some $\gamma$-approximation matrix on $(C,k)$. Then, we can estimate the upper bound of the clustering error $\|A - \tilde{X}_{opt}\tilde{X}_{opt}^T A\|_F^2$ as follows (Boutsidis and Magdon-Ismail 2013).

$$\|A - \tilde{X}_{opt}\tilde{X}_{opt}^T A\|_F^2 = \|(I_m - \tilde{X}_{opt}\tilde{X}_{opt}^T)AZZ^T + (I_m - \tilde{X}_{opt}\tilde{X}_{opt}^T)E\|_F^2. \tag{19}$$

Because $Z^T E^T = 0_{k \times m}$ we have

$$((I_m - \tilde{X}_{opt}\tilde{X}_{opt}^T)AZZ^T)((I_m - \tilde{X}_{opt}\tilde{X}_{opt}^T)E)^T = 0_{m \times m}.$$

Consequently, equation (19) can be reduced to

$$\begin{aligned} \|A - \tilde{X}_{opt}\tilde{X}_{opt}^T A\|_F^2 &= \|(I_m - \tilde{X}_{opt}\tilde{X}_{opt}^T)AZZ^T\|_F^2 + \|(I_m - \tilde{X}_{opt}\tilde{X}_{opt}^T)E\|_F^2 \\ &\leq \|(I_m - \tilde{X}_{opt}\tilde{X}_{opt}^T)AZZ^T\|_F^2 + \|E\|_F^2 \end{aligned} \quad (20)$$

Given $\Omega$ and $S$, we have (Boulsidis and Margdon-Ismail, 2013)

$$AZZ^T = A\Omega S(Z^T\Omega S)^+ Z^T + Y, \quad (21)$$

where $Y \in R^{m \times n}$ is a residual matrix and $(.)^+$ denotes the pseudo-inverse of a matrix. It is noted that

$$\|AB\|_F \leq \|A\|_F \|B\|_F, \quad \|WZ^T\|_F = \sqrt{\text{Tr}(WZ^TZW)} = \|W\|_F \text{ and for any two matrices,}$$

$$\|Y_1 + Y_2\|_F^2 \leq 2\|Y_1\|_F^2 + 2\|Y_2\|_F^2.$$

Then, the first term in equation (20) can be further bounded by

$$\begin{aligned} \|(I_m - \tilde{X}_{opt}\tilde{X}_{opt}^T)AZZ^T\|_F^2 &\leq 2\|(I_m - \tilde{X}_{opt}\tilde{X}_{opt}^T)A\Omega S(Z^T\Omega S)^+ Z^T\|_F^2 + 2\|Y\|_F^2 \\ &\leq 2\|(I_m - \tilde{X}_{opt}\tilde{X}_{opt}^T)A\Omega S\|_F \|(Z^T\Omega S)^+\|_F + 2\|Y\|_F \end{aligned} \quad (22)$$

Using equation (17), we obtain

$$\begin{aligned} \|(I_m - \tilde{X}_{opt}\tilde{X}_{opt}^T)AZZ^T\|_F &\leq 2\gamma \|(I_m - X_{opt}X_{opt}^T)A\Omega S\|_F^2 \|(Z^T\Omega S)^+\|_F^2 + 2\|Y\|_F^2 \\ &\leq 2\gamma \frac{|(I_m - X_{opt}X_{opt}^T)A\Omega S\|_F^2}{\sigma_k^2(Z^T\Omega S)} + 2\|Y\|_F^2 \end{aligned} \quad (23)$$

Since rank $(Z^T\Omega S) = k$, we have

$$Z^T\Omega S(Z^T\Omega S)^+ = I_k$$

and

$$AZZ^T - AZZ^T\Omega S(Z^T\Omega S)^+ Z^T = 0_{m \times n}, \text{ which implies that}$$

$$\begin{aligned} Y &= AZZ^T - A\Omega S(Z^T\Omega S)^+ Z^T \\ &= AZZ^T - AZZ^T\Omega S(Z^T\Omega S)^+ Z^T - (A - AZZ^T)\Omega S(Z^T\Omega S)^+ Z^T \\ &= -(A - AZZ^T)\Omega S(Z^T\Omega S)^+ Z^T. \end{aligned}$$

Therefore, we have

$$\begin{aligned}
\|Y\|_F^2 &= \|(A - AZZ^T)\Omega S(Z^T\Omega S)^+ Z^T\|_F^2 \\
&\leq \|(A - AZZ^T)\Omega S\|_F^2 \|(Z^T\Omega S)^+ Z^T\|_F^2 \\
&\leq \|(A - AZZ^T)\Omega S\|_F^2 \|(Z^T\Omega S)^+\|_F^2 \\
&= \frac{\|(A - AZZ^T)\Omega S\|_F^2}{\sigma_k^2(Z^T\Omega S)}.
\end{aligned} \qquad (24)$$

Combining equations (23) and (24), we obtain:

$$\begin{aligned}
\|(I_m - \widetilde{X}_{opt}\widetilde{X}_{opt}^T)AZZ^T\|_F^2 &\leq 2\gamma \frac{|(I_m - X_{opt}X_{opt}^T)A\Omega S\|_F^2}{\sigma_k^2(Z^T\Omega S)} + 2\|Y\|_F^2 \\
&\leq 2\frac{\gamma|(I_m - X_{opt}X_{opt}^T)A\Omega S\|_F^2 + 2\|(A - AZZ^T)\Omega S\|_F^2}{\sigma_k^2(Z^T\Omega S)} \\
&\leq 2\frac{\gamma\|(I_m - X_{opt}X_{opt}^T)A\Omega S\|_F^2 + \|E\Omega S\|_F^2}{\sigma_k^2(Z^T\Omega S)}
\end{aligned} \qquad (25)$$

Combining equations (20) and (25) we obtain the following upper bound:

$$\|A - \widetilde{X}_{opt}\widetilde{X}_{opt}^T A\|_F^2 \leq 2\frac{\gamma|(I_m - X_{opt}X_{opt}^T)A\Omega S\|_F^2 + \|E\Omega S\|_F^2}{\sigma_k^2(Z^T\Omega S)} + \|E\|_F^2. \qquad (26)$$

The upper bound provides information about how to choose $Z$, $\Omega$ and $S$. We chose $Z$ to make the residual $E$ small. Several terms in the upper bound can be used to guide the selection of the sampling and rescaling matrices $\Omega$ and $S$. The first term in the numerator of the upper bound is the clustering error of the input partition in the reduced dimension space. We chose $\Omega$ and $S$ to make this clustering error small. The residual $E$ is involved in the second term of the numerator and final term in the inequality (26). We chose $\Omega$ and $S$ such that they will not substantially increase the size of the residual $E$. The term in the denominator involves $Z, \Omega$ and $S$. Therefore, the selected $\Omega$ and $S$ do not significantly change the singular structure of the projection matrix $Z$ and ensure that $\sigma_k^2(Z^T\Omega S)$ is large. Under these guidance, the following randomized feature selection algorithm can be developed.

## 2.4 Randomized feature selection algorithms

Let $k$ be the number of clusters and $\varepsilon$ be the errors that are allowed.

Set $r = k + \left\lceil \dfrac{k}{\varepsilon} \right\rceil + 1$ as the number of features being selected. Consider data matrix

$$A = \begin{bmatrix} a_{11} & \cdots & a_{1n} \\ \vdots & \cdots & \vdots \\ a_{m1} & \cdots & a_{mn} \end{bmatrix}.$$

Let $i$ denote the index of the individual sample and $j$ be the index of feature. We intend to select $r$ features.

Procedures of algorithms are given as follows.

1. Generate an $n \times r$ standard Gaussian matrix R, with $R_{ij} \sim N(0,1)$.

2. Let $Y = AR \in R^{m \times r}$.

3. Orthonormalize the columns of the matrix $Y$, which leads to the matrix $Q \in R^{m \times r}$.

4. Singular value decomposition of the matrix $Q^T A$: $Q^T A = U\Sigma V^T$.

Let $Z \in R^{n \times k}$ be the top $k$ right singular vectors of $Q^T A$, i.e., $Z = [V_1,...,V_k]$.

5. Calculate the sampling probability:

$P_i = \dfrac{\|Z_{(i)}\|_2^2}{\|Z\|_F^2}, i = 1,...,n, \sum_{i=1}^{n} P_i = 1$, where $Z_{(i)}$ is the $i$-th row of the matrix $Z$ and

$\|Z\|_F^2 = tr(ZZ^T)$.

6. Initiate $\Omega = 0_{n \times r}$ and $S = 0_{r \times r}$.

For $t = 1,...,r$, pick an integer $i_t$ from the set $\{1,2,...,n\}$ with probability $P_{i_t}$ and replacement, set

$\Omega(i_t, t) = 1$ and $S(t,t) = \dfrac{1}{\sqrt{rP_{i_t}}}$.

End

7. Return $C = A\Omega S \in R^{m \times r}$.

## 3. Results

We tested our algorithm on two cancer histology image datasets downloaded from the TCGA database. The first dataset is an ovarian cancer dataset, which includes 176 histology images taken from 106 drug sensitive and 76 drug resistant tissue samples. The second dataset is Kidney Renal Clear Cell Carcinoma (KIRC) dataset which includes 188 histology images taken from 121 KIRC tumor and 67 normal tissue samples.

We compared the performance of our algorithm with the standard $K$ - means and regularization-based sparse $K$ - means clustering algorithms (Witten and Tibshirani 2010). We also compared the performance of the two FPCA with the Fourier expansion and SIFT descriptor (Chen and Tian 2006  ). We use cluster accuracy (ACC) which is defined as the proportion of correctly clustered images, cluster sensitivity which is defined as the proportion of correctly clustered drug sensitive or tumor samples, and cluster specificity which is defined as the proportion of correctly clustered drug resistant or normal samples, for performance evaluation in this study..

### *3.1. Comparison of two dimensional FPCA with Fourier expansion and SIFT descriptor*

To intuitively illustrate the power of FPCA to reduce the dimension of image data, we first presented Figure 1 which showed the original and reconstructed lena face and KIRC tumor cell images. We observed that the reconstructed lena and images of KIRC tumor cell using only 6 functional principal components (FPCs) and 188 FPCs, respectively, are very close to the original images. However, even when we used the16, 129 terms in the Fourier expansion to reconstruct the lena face and KIRC cell images, the reconstructed images were still unclear. Then, we compared the accuracies of the standard k-means algorithms for clustering ovarian cancer and KIRC tissue samples using FPC scores, Fourier expansion coefficients and SIFT descriptors as image features. The results were summarized in Table 1. We observed from Table 1 that the cluster analysis using FPC scores as features has a higher accuracy than using Fourier expansion coefficients and SIFT descriptors for both ovarian cancer and KIRC datasets.

### 3.2. Performance of standard k-means clustering algorithm, sparse k-means clustering algorithm and randomized sparse k-means clustering algorithm

We compared the performance of the standard *k*-means clustering algorithm and sparse *k*-means clustering algorithm and randomized sparse *k*-means clustering algorithm in the ovarian and KIRC cancer studies. The SPARCL package was used for implementing the *sparse K -means clustering algorithm* (Witten and Tibshirani 2010). The SIFT descriptor (Lowe 2004) was used as features. The images in the ovarian cancer study were taken before treatment. Therefore, the images were used to predict drug response. The results were summarized in Table 2. Table 2 showed that the randomized k-means clustering algorithms used significantly smaller features, but achieved higher accuracy than both the standard K-means and sparse k-means algorithms.

### 3.3. Performance of standard k-means, sparse k-means and randomized sparse k-means clustering algorithms using FPC scores

We studied the performance of standard k-means, sparse k-means and randomized sparse k-means clustering algorithm using the FPC scores as image features. The results of this application of three clustering algorithms to two cancer imaging datasets were summarized in Table 3. Again, the randomized sparse k-means algorithms used the smallest number of FPC scores, but had the highest clustering accuracy, followed by sparse k-means clustering algorithms. The standard k-means clustering algorithms used the largest number of FPC scores, but achieved the lowest clustering accuracy. Comparing Table 3 with Table 2, we found that FPCA substantially improved clustering accuracy. Specifically, for the KIRC dataset we observed that replacing the SIFT descriptor with FPC scores increased the clustering accuracies of the stand k-means, sparse k-means and randomized sparse k-means from 68.09% to 80.85%, 58.51% to 81.91%, and 72.87% to 83.51%, respectively.

### *3.4. Performance of standard spectral, sparse K-means, and randomized sparse spectral clustering algorithms using Fourier expansion coefficients*

To further evaluate the performance of randomized sparse clustering algorithm, we considered spectral clustering algorithm, another type of clustering algorithms (Liu et al., 2013). We used three algorithms: standard spectral, sparse k-means and randomized spectral clustering algorithms with Fourier expansion coefficients to conduct clustering analysis for the ovarian cancer and KIRC datasets. Table 4 was presented to summarize the results. The performance patterns of the three clustering algorithms using Fourier expansion coefficients as imaging features were the same as that using other features. Table 4 showed that sparse principle for spectral clustering algorithms still improved cluster accuracy and randomized sparse clustering algorithms had the highest accuracy among the three clustering algorithms. We also observed that in general, using Fourier expansion coefficients as imaging features had less accuracy than using FPC scores as features (See Tables 4 and 5).

### *3.5. Performance of standard spectral, sparse K-means, and randomized sparse spectral clustering algorithms using FPC score*

FPCA can improve clustering accuracy. This does not depended on which clustering algorithms are used. Table 5 showed the performance of standard spectral, sparse k-means and randomized spectral clustering algorithms with FPC scores in cluster analysis of two TCGA cancer datasets. We can clearly see that standard spectral, sparse k-means and randomized spectral clustering algorithms with FPC scores took much less features, but can reach as high as or even better clustering accuracy than these methods using Fourier expansion coefficients as features. Table 5 again demonstrated that in most cases FPCA can substantially improve the performance of the clustering algorithms.

### *3.6. Multiple cluster analysis*

Although a population can be divided into two groups: normal and patient groups, in general, patients' subpopulation is highly heterogeneous and has complex structures. Patients need to be further divided into several more homogeneous groups. Table 6 presented results of three clustering algorithms for multiple cluster analysis in the KIRC studies where tumor cells were partitioned into three groups. Neoplasm histologic grade which is based on the microscopic morphology of a neoplasm with hematoxylin and eosin (H&E) staining (G1, G2, G3 and G4) was selected as prognostic factors of survival (Erdogan et al., 2004). In the present analysis, patients of G1 and G2 were regrouped as group 1 patients. Patients of G3 were regrouped as group 2 patients and patients of G4 were regrouped as group 3 patients. Table 6 suggested that the randomized sparse k-means had the highest accuracy for clustering KIRC tumor cell grades, followed by sparse k-means and standard k-means clustering algorithms, where accuracy was defined as the proportion of individuals who were correctly assigned to groups. As shown in Figure 2, clustering tumor cells has a close relationship with cell pathology which characterizes progressing and development of tumors. In Figure 2a, morphology of nucleus that was represented by black circles changed slowly. When disease proceeded nucleus became large and expanded (Figure 2b). When tumors proceeded to the final stage, the nucleus was metastated and became blur (Figure 2c).

### *4. Discussion*

In this paper, we proposed to combine feature extraction and feature selection for cluster analysis of imaging data and developed FPCA-based randomized sparse clustering algorithms. The data in image applications are high-dimensional. Dimension reduction is a key to the success of imaging cluster analysis. To successfully perform image cluster analysis, we addressed several issues for dimensional reduction in the sparse image cluster analysis.

First issue is how to use feature extraction to reduce the data dimension. In other words, we construct a small set of new artificial features that are often linear combinations of the original features and then the k-means method is used to cluster on the constructed features. A variety of methods for feature extraction has been developed. PCA or FPCA are popular methods for feature extraction. However, FPCA is developed for one dimensional data and cannot be simply applied to two or three dimensional imaging data.  Here we extended FPCA from one dimension to two or three dimensions and applied it to extraction of imaging features. Real histology imaging cluster analysis showed that the FPCA for imaging dimension reduction substantially outperformed the SIFT descriptor and Fourier expansion and is the one of choice for imaging feature extraction.

A second issue is to develop sparse clustering algorithms that attempts to identify features underlying clusters and remove noise and irrelevant variables. There are two types of sparse clustering algorithms. One type of algorithms is to optimize weighted within-cluster sum of squares and use a lasso type penalty to select weights and features. The difficulty with this type of constrained based sparse clustering algorithms is how to determine a threshold that is used to remove features. In theory, the features corresponding to non-zero weights will be selected for clustering. In practice, all weights vary continuously. Determining an appropriate threshold to cut of irrelevant features is a difficult challenge. An alternative approach is to randomly and directly select a small subset of the actual features that can ensure to approximately reach the optimal k-means objective value. Both mathematic formulations of the k-means objective function and sampling algorithms to optimize objective function have well be developed. We can expect that the developed randomized sparse k-means clustering algorithms can work very well. Using real cancer imaging data, we showed that (1) both randomized k-means clustering and lasso-type k-means clustering algorithms substantially outperformed the standard k-means algorithm, and (2) performance of the randomized k-means sparse clustering algorithm was much better than that of the lasso type sparse k-means clustering algorithms.

A third issue is to combine feature extraction and feature selection. Feature extraction and feature selection are two major tools for dimension reduction. In imaging cluster analysis, feature extraction and feature selection are often used separately for data reduction. The main strength of our approach is to integrate feature extraction and feature selection into a dimension reduction tool before clustering images. We first performed two dimensional FPCA of images to extract group structure information of images. The resulting vector of FPC scores containing image group information were used to represent the features of the images. Then, we designed a random matrix column selection algorithm to select some components of the vector of FPC scores for further cluster analysis. Finally, k-means method was used to cluster the selected FPC scores. We showed that k-means method with combined feature extraction and feature selection as dimension reduction had the highest cluster accuracy in two real cancer clustering studies.

The fourth issue is the independence of dimension reduction benefits for clustering from the used clustering methods. Appropriate use of feature extraction and feature reduction may substantially improve the performance of clustering algorithms. This conclusion will not depend on which clustering algorithms are used. We demonstrated that cluster accuracies of both sparse k-means and sparse spectral clustering were higher than standard k-means and spectral clustering without dimension reduction.

While the proposed method provides a powerful approach to image cluster analysis, some challenges still remain. The randomized feature selection algorithms have deep connections with the objective function of $k$-means clustering and low-rank approximations to the data or feature matrix. However, the solutions to optimize the objective function of $k$-means clustering may not correspond to the true group structure of the image data well, which in turn, will compromise the performance of the randomized feature selection methods. Selection of the number of features also depends on the accuracy of low-rank approximation. Although we can provide theoretic calculation of the number of selected features, in practice we need to automatically calculate it by iterating the feature selection algorithm from the data, which requires heavy computation for large datasets. The randomized feature selection for multiple groups clustering still has serious limitation. Clustering images into multiple groups is an important, but a challenge problem. The main purpose of this paper is to stimulate discussion about what are the optimal strategies for high dimensional image cluster analysis. We hope that our results will greatly increase confidence in applying the dimension reduction to image cluster analysis.

## 5. Conclusions

We extended one dimensional FPCA to the two dimensional FPCA and develop novel sparse cluster analysis methods which combine two dimensional FPCA with randomized feature selection to reduce the high dimension of imaging data. We used stochastic calculus to derive the formula for calculation of the variance of integral of weighted linear combination of two dimensional signals of images. We formulated two dimensional FPCA as a maximization of this variance with respect to weight function (functional components) of two variants and used variation of theory to find solutions that are solutions to integral equations with two variants. We used functional expansion to develop computational methods for solving integral equations with respect to functional components and finding FPC scores which are taken as features for cluster analysis.

Followed the approach of Boutsidis et al. (2013), we explored matrix approximation theory and a technique of Rudelson and Vershynin (2007) to design a randomized method to select FPC scores as features for cluster analysis with probability that are correlated with the right singular vectors of the FPC score matrix. In theory, we can prove that the randomized feature selection algorithm guarantees the quality of the resulting clusters. The developed randomized algorithms integrating FPC scores as features for dimension reduction can be applied to k-means and spectral clustering algorithms. Results on clustering histology images in the ovarian cancer and KIRC cancer studies showed that the randomized *k*-means and spectral clustering algorithms integrating FPCA substantially outperform other existing clustering algorithms with and without feature selection. The randomized sparse clustering algorithms integrating FPCA is a choice of methods for image clustering.

*Acknowledgments*

The project described was supported by grants 1R01AR057120–01 and 1R01HL106034-01 from the National Institutes of Health and NHLBI, respectively. The authors wish to acknowledge the contributions of the research institutions, study investigators, field staff and study participants in creating the TCGA datasets for biomedical research.

*Conflict of interest*

The authors declare that they have no competing interests.

*Vitae*

Nan Lin received four year undergraduate education at the Jilin University in China and got the bachelor degree in biomedical engineering. Then he continued his graduate study in the field of biomedical engineering and got the master degree in biomedical engineering at Purdue University. After that, he moved to the University of Illinois, Urbana-Champaign and received two years' graduate training in statistics and got the master degree in statistics. Currently, I am pursuing my Ph.D degree in biostatistics at the University of Texas, Health Science Center at Houston.

Junhai Jiang is a Ph.D. student at the University of Texas Health Science Center at Houston. He got his Bachelor's degree at the Fudan University in 2011 and his major was statistics. Afterwards, he got the degree of Master of Science in the Department of Mathematics and Statistics in Texas Tech University in 2013, and right now, he is a biostatistic Ph. D  student under Dr. Momiao Xiong's supervision.

Shicheng Guo is a Ph.D. candidate student in medical genetics program, supervised by Dr. Li Jin, at the Fudan University, Shanghai, China. He obtained his bachelor degree from the school of life sciences, Northeast Agricultural University, Harbin, China in 2009. Currently, he is a visiting student and research assistant at Dr. Momiao Xiong's lab in the School of public health, University of Texas Health Science Center at Houston. He is focusing on cancer related susceptibility variants identification, diagnostic and prognostic biomarker discovery and risk prediction based on genetic and epigenetic variations as well as cancer early diagnosis based on the joint analysis of high-dimensional image, RNA-seq, methylation-seq datasets.

Momiao Xiong, Ph. D, Professor, Division of Biostatistics and Human Genetics Center, School of Public Health, the University of Texas Health Science Center at Houston . He obtained his PhD degree from the Department of Statistics at the University of Georgia in 1993 and finished his postdoctoral training in computational biology from the University of Southern California in 1995. His research is in the area of statistical genetics, bioinformatics, machine learning and image analysis.

*Figure Legend*

**Figure 1.** (a)**.** Original image of the Lena face image. (b) Six lena face images are downloaded from the USC SIPI Image Database (Web: http://sipi.usc.edu/database/). All six lena face images were used to perform FPCA.  Reconstruction of the images of one of the six lena face images using its 6 FPC scores. (c) All six lena face images were used to perform the Fourier expansion. Reconstruction of the  same lena face  image as that in (b) using its  first 16129 Fourier coefficients (d) Original image of one of the 121 histology images of  the kidney cancer cells.  (e) A total of 188 kidney histology images (121 kidney cancer cells and 67 kidney normal cells) were downloaded from the TCGA database.  All images were used to perform FPCA.  Reconstruction of histology images of kidney cancer cells by using its 188 FPCA scores.   (f)  All 188 kidney histology images were used to perform the Fourier expansion. Reconstruction of the same kidney histology image as that in (e) by using its first 16,129 Fourier expansion coefficients

**Figure 2**. Historic pathology images. (a) Pathology grades 1 and 2, (b) pathology grade 3 and (c) pathology grade 4.

## Tables

Table 1. Performance of standard *k*-means clustering algorithm for FPCA, descriptor and Fourier expansion

|            | Ovarian Cancer |             |             | KIRC     |             |             |
|------------|----------|-------------|-------------|----------|-------------|-------------|
|            | Accuracy | Sensitivity | Specificity | Accuracy | Sensitivity | Specificity |
| FPCA       | 0.570    | 0.660       | 0.400       | 0.835    | 0.926       | 0.672       |
| Descriptor | 0.557    | 0.547       | 0.547       | 0.681    | 0.587       | 0.701       |
| Fourier    | 0.557    | 0.557       | 0.557       | 0.803    | 0.917       | 0.597       |

Table 2. Performance of standard K-means, sparse K-means and randomized K-means clustering algorithm using the SIFT descriptor

|  | Ovarian Cancer | | | | KIRC | | | |
|---|---|---|---|---|---|---|---|---|
|  | Features | Accuracy | Sensitivity | Specificity | Features | Accuracy | Sensitivity | Specificity |
| *k*-means | 2,560 | 0.557 | 0.547 | 0.547 | 2,560 | 0.681 | 0.587 | 0.701 |
| Sparse *k*-means | 574 | 0.545 | 0.472 | 0.657 | 597 | 0.585 | 0.62 | 0.522 |
| Randomized *k*-means | 70 | 0.608 | 0.708 | 0.457 | 100 | 0.729 | 0.818 | 0.567 |

Table 3. Performance of standard *k*-means, sparse *k*-means and randomized sparse *k*-means clustering algorithms using FPC scores.

|  | Ovarian Cancer | | | | KIRC | | | |
|---|---|---|---|---|---|---|---|---|
|  | Features | Accuracy | Sensitivity | Specificity | Features | Accuracy | Sensitivity | Specificity |
| *k*-means | 176 | 0.574 | 0.660 | 0.400 | 188 | 0.809 | 0.917 | 0.612 |
| Sparse *k*-means | 81 | 0.585 | 0.670 | 0.457 | 92 | 0.819 | 0.819 | 0.642 |
| Randomized sparse *k*-means | 23 | 0.653 | 0.793 | 0.486 | 5 | 0.835 | 0.926 | 0.672 |

Table 4. Performance of standard spectral, sparse *k*-means clustering and sparse spectral with randomized feature selection clustering algorithms with Fourier expansion.

| | ovarian cancer | | | | KIRC | | | |
|---|---|---|---|---|---|---|---|---|
| | Features | Accuracy | Sensitivity | Specificity | Features | Accuracy | Sensitivity | Specificity |
| Spectral clustering | 65025 | 0.5568 | 0.5566 | 0.5571 | 65025 | 0.8032 | 0.9174 | 0.5970 |
| Sparse *k*-means | 959 | 0.5455 | 0.5000 | 0.6143 | 161 | 0.8191 | 0.9174 | 0.6418 |
| Randomized Spectral clustering | 100 | 0.6420 | 0.5755 | 0.7429 | 10 | 0.8351 | 0.9256 | 0.6716 |

Table 5. Performance of standard spectral, sparse *k*-means, and sparse spectral with randomized feature selection clustering algorithms using FPC scores as features

|  | ovarian cancer | | | | KIRC | | | |
| --- | --- | --- | --- | --- | --- | --- | --- | --- |
|  | Features | Accuracy | Sensitivity | Specificity | Features | Accuracy | Sensitivity | Specificity |
| Spectral clustering | 176 | 0.585 | 0.670 | 0.460 | 188 | 0.819 | 0.917 | 0.642 |
| Sparse *k*-means | 176 | 0.545 | 0.500 | 0.614 | 188 | 0.819 | 0.860 | 0.746 |
| Randomized Spectral clustering | 23 | 0.688 | 0.858 | 0.429 | 13 | 0.835 | 0.909 | 0.702 |

Table 6. Performance of standard *k*-means, sparse *k*-means and randomized *k*-means algorithms for clustering KIRC tumor cell grades

|  |  | TRUE | | |
|---|---|---|---|---|
| Method | Assigned | Group1 | Group 2 | Group 3 |
| *k*-means | Group 1 | 17 (58.6%) | 15 (53.6%) | 7 (50.0%) |
|  | Group 2 | 12 (41.4%) | 12 (42.9%) | 7 (50.0%) |
|  | Group 3 | 0 | 1 (3.4%) | 0 |
|  | Accuracy | 40.80% | | |
| Sparse *k*-means | Group 1 | 10 (34.5%) | 6 (21.4%) | 3 (21.4%) |
|  | Group 2 | 13 (44.8%) | 17 (60.7%) | 7(50.0%) |
|  | Group 3 | 6 (20.7%) | 5 (17.9%) | 4 (28.6%) |
|  | Accuracy | 43.70% | | |
| Randomized sparse *k*-means | Group 1 | 14 (48.3%) | 4 (14.3%) | 2 (14.3%) |
|  | Group 2 | 8 (27.6%) | 20 (71.4%) | 8 (57.1%) |
|  | Group 3 | 7 (24.1%) | 4 (14.3%) | 4 (28.6%) |
|  | Accuracy | 53.50% | | |

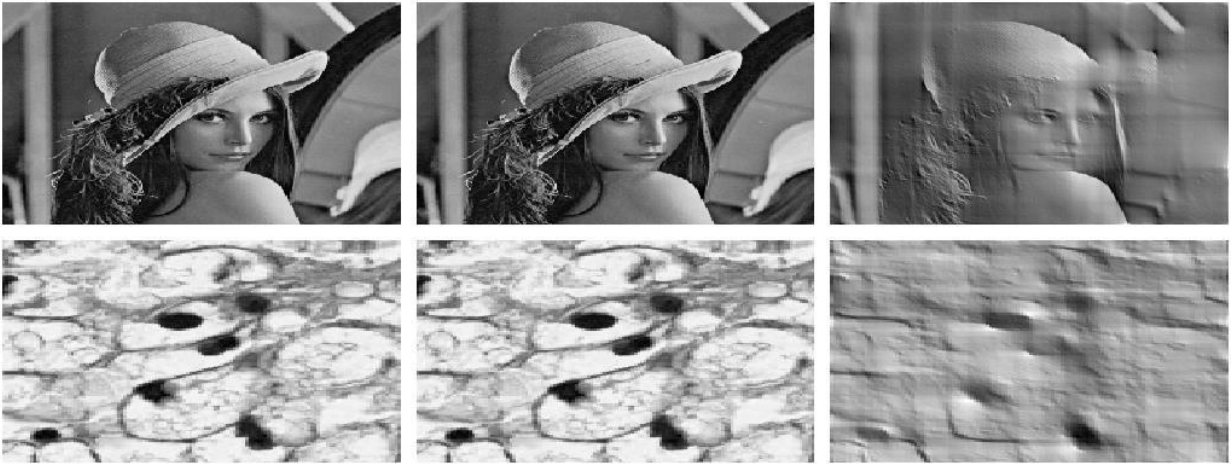

Figure 1.

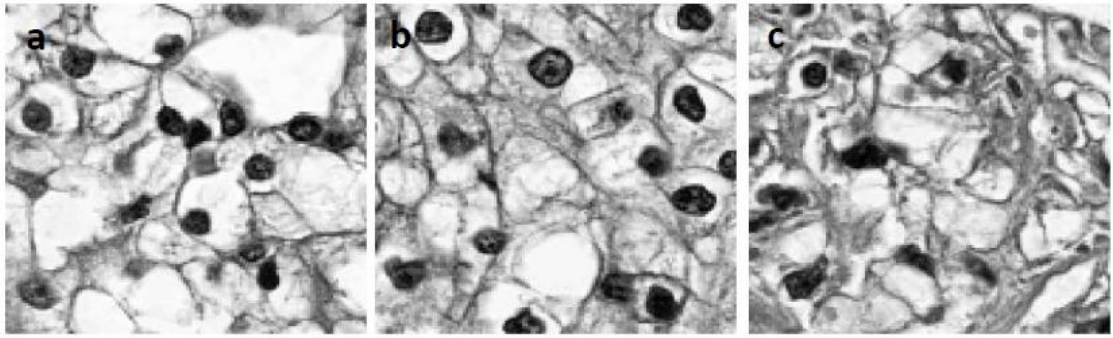

Figure 2.